\pdfoutput=1
\documentclass[letterpaper, 10 pt, conference]{ieeeconf}  % Comment this line out if you need a4paper

\usepackage{blindtext, graphicx}
\usepackage{gensymb}
\usepackage{xcolor}

\usepackage{tikz,lipsum}
\usepackage[labelformat=simple]{subcaption}

\DeclareCaptionLabelSeparator{periodspace}{.\quad}
\usepackage{listings}

\usepackage{algorithm}
\usepackage{algorithmic}

\usepackage{framed} 

\IEEEoverridecommandlockouts                              % This command is only needed if 
\title{\LARGE \bf
Automatic Calibration of a Robot Manipulator\\ and Multi 3D Camera System
}

\author{Justinas Mi\v{s}eikis$^{1}$, Kyrre Glette$^{2}$, Ole Jakob Elle$^{3}$, Jim Torresen$^{4}$% <-this % stops a space
\thanks{$^{1}$ $^{2}$ $^{3}$ $^{4}$Justinas Mi\v{s}eikis, Kyrre Glette, Ole Jakob Elle and Jim Torresen are with the Department of Informatics, University of Oslo, Oslo, Norway}
\thanks{$^{1}$ $^{2}$ $^{4}$ {\tt\small \{justinm,kyrrehg,jimtoer\}@ifi.uio.no}}%
\thanks{$^{3}$Ole Jakob Elle has his main affiliation with The Intervention Centre, Oslo University Hospital, Oslo, Norway
        {\tt\small oelle@ous-hf.no}}%
}

\begin{document}

%The sharing worked! Kyrre

\maketitle
\thispagestyle{empty}
\pagestyle{empty}

%%%%%%%%%%%%%%%%%%%%%%%%%%%%%%%%%%%%%%%%%%%%%%%%%%%%%%%%%%%%%%%%%%%%%%%%%%%%%%%%
\begin{abstract}

With 3D sensing becoming cheaper, environment-aware and visually-guided robot arms capable of safely working in collaboration with humans will become common. However, a reliable calibration is needed, both for camera internal calibration, as well as Eye-to-Hand calibration, to make sure the whole system functions correctly. We present a framework, using a novel combination of well proven methods, allowing a quick automatic calibration for the integration of systems consisting of the robot and a varying number of 3D cameras by using a standard checkerboard calibration grid. Our approach allows a quick camera-to-robot recalibration after any changes to the setup, for example when cameras or robot have been repositioned. Modular design of the system ensures flexibility regarding a number of sensors used as well as different hardware choices. The framework has been proven to work by practical experiments to analyze the quality of the calibration versus the number of positions of the checkerboard used for each of the calibration procedures.

\end{abstract}

%%%%%%%%%%%%%%%%%%%%%%%%%%%%%%%%%%%%%%%%%%%%%%%%%%%%%%%%%%%%%%%%%%%%%%%%%%%%%%%%
\section{INTRODUCTION}

In many practical applications, industrial robots are still working "blind" with hard-coded trajectories. This results in the workspace for robots and humans being strictly divided in order to avoid any accidents, which, unfortunately, sometimes still occur. Furthermore, working in a dynamic environment without a direct connection to other machinery sharing the same workspace, might prove difficult. It is often more common to have collision detection systems, which do not always work as expected, rather than collision prevention methods~\cite{ur5collision}. However, \textit{environment-aware robots}~\cite{flacco2012depth}~\cite{rakprayoon2011kinect} are becoming more common, both developed in research and by robot manufacturers themselves, e.g. Baxter by Rethink Robotics~\cite{fitzgerald2013developing}. 

\begin{figure}[h]
\centering
\includegraphics[width=0.40\textwidth]{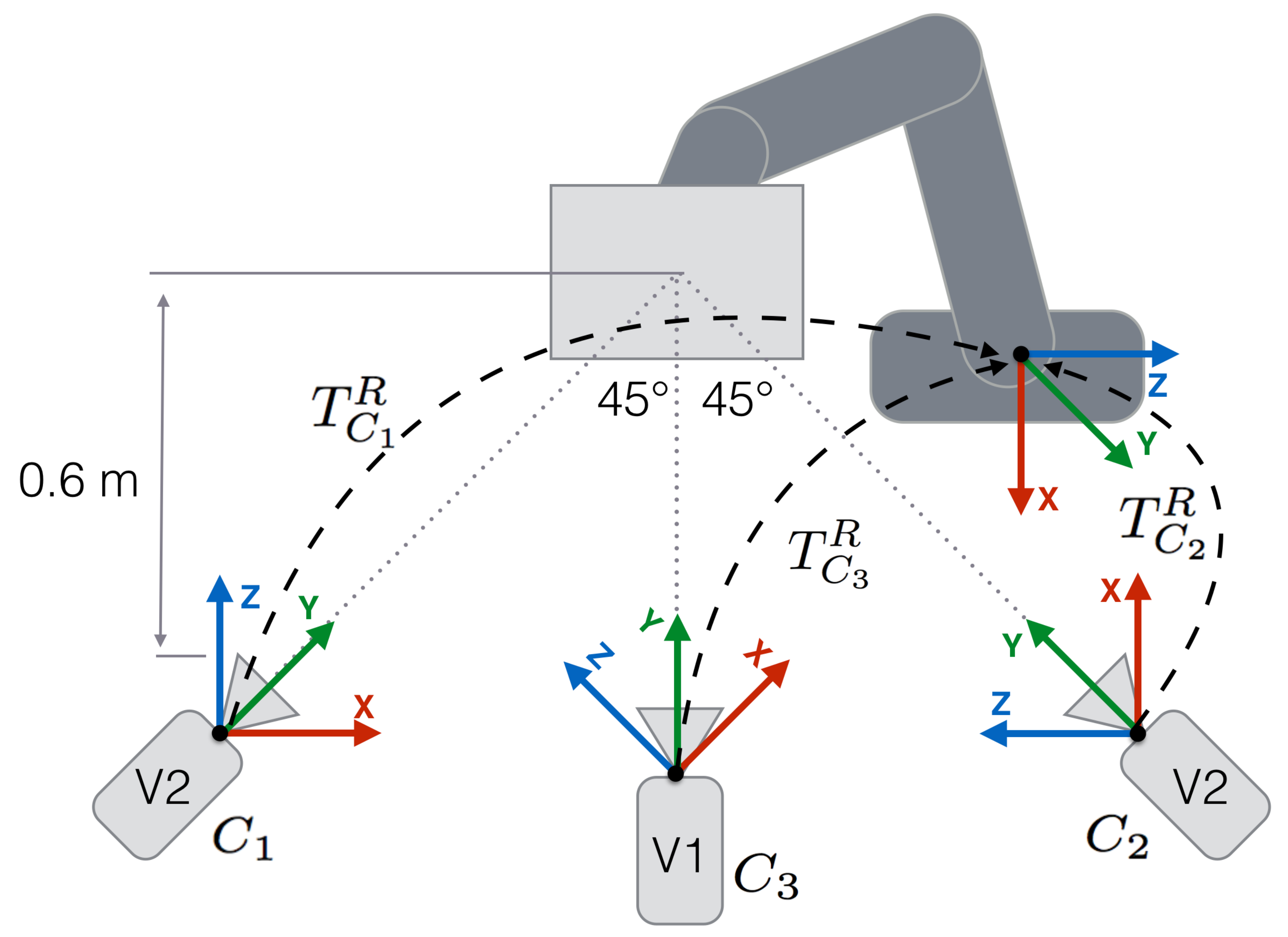}
\caption{System Setup with two Kinect V2 depth sensors aimed at the robot end effector at approximately 45{\degree}  viewpoints and a Kinect V1 sensor placed between them facing the robot. In the system, Eye-to-Hand calibration is represented by the Affine transformation matrix $T_{C}^{R}$, which transforms the coordinate system of each camera to the coordinate system of the robot base, making it common for the whole setup.}
\label{fig:system_setup}
\vspace{-0.2cm}
\end{figure}

Low-cost and high-accuracy 3D cameras, also called RGB-D sensors, like Kinect V1 and V2~\cite{Fankhauser2015KinectV2ForMobileRobotNavigation}, are already available. They are suitable for a precise environment sensing in the workspace of a robot, providing both color image and depth information~\cite{smisek20133d}. However, external sensors are commonly used in \textit{fixed positions} around the robot and are normally not allowed to be moved. After any reconfiguration in the setup, the whole system has to be calibrated, usually by a skilled engineer. Camera calibration can be divided into two main stages:
\begin{itemize}
\item Internal camera parameters, like lens distortion, focal length, optical center, and for RGB-D cameras, color and depth image offsets~\cite{opencvchessboard}~\cite{Fankhauser2015KinectV2ForMobileRobotNavigation}.
\item External camera parameters: the pose (position and orientation) of a camera in a reference coordinate frame. It is commonly called Eye-to-Hand calibration~\cite{ma2014hand}~\cite{horaud1995hand}. The Eye-to-Hand calibration, or transformation from the camera coordinate system to the robot base coordinate system is shown in Figure~\ref{fig:system_setup}.
\end{itemize}

Normally, it is sufficient to perform an internal camera parameter calibration only once per sensor unless the lens or sensor itself will be changed or modified. Reliable calibration methods already exist, which are widely used~\cite{tsai1987versatile}~\cite{iaikinect2}~\cite{foix2011lock}~\cite{amon2014evaluation}.

Eye-to-Hand calibration, on the other hand, is more application specific and crucial for precise environment sensing by the robot or vision guided robot control (visual servoing)~\cite{lippiello2005eye}. Some work has been successful in calibrating multiple cameras and a robot using a custom-made target object placed in a common field of view for all the sensors in the workspace~\cite{heikkila2000flexible}. Another method calibrated multiple cameras fixed on a rig using structure-of-motion method to estimate relative positions between the cameras~\cite{esquivel2007calibration}. A similar approach was used for calibrating a network of Kinect sensors aimed at robotic inspection of large work-spaces, where sensors are in fixed positions~\cite{macknojia2013calibration}. Robot arm mounted camera, also known as Eye-in-Hand, calibration by moving it to the points of the calibration grid, which is in a fixed position was also proposed~\cite{zhuang1995simultaneous}~\cite{dornaika1998simultaneous}. However, most of the presented work is either aimed at the very specific setup or requires a large amount of manual placement of calibration grids, making it time-consuming.

This paper presents a framework to be used for an automatic combined internal camera parameter and Eye-to-Hand calibration by utilizing a robot arm manipulator to actively move around a standard checkerboard calibration grid. The framework is using existing and reliable calibration approaches, but is based on a novel combination of methods to make the calibration process fully automatic and adaptable to as few or as many external 3D cameras as needed. Furthermore, an end-effector to the checkerboard offset is estimated, so a variety of end-effector attachments can be used. It is a time saving and flexible process without requiring any additional equipment for preparing the setup, just a slightly modified A4 size printed checkerboard.

The whole framework is based on the Robot Operating System (ROS) and making use of the modular design and available integration for a large amount of robot and sensor types~\cite{quigley2009ros}. Each part of the algorithm is split into a number of separate modules communicating in between each other using pre-defined message formats. The benefits of this approach is the ability to easily modify parts of the process without affecting the rest of processing as well as to include additional processing steps if needed. Furthermore, each framework module can be reused given that the input and output inter-module message format matches.

This allows the actual hardware, robot and 3D cameras, to be interchangeable by simply modifying the configuration file, as long as they have ROS-supported drivers. Only minimal supervision is required during the whole process.

This paper is organized as follows. We present the system setup in Section~\ref{sec:system_setup}. Then, we explain the method in Section~\ref{sec:method}. We provide experimental results in Section~\ref{sec:experiments}, followed by relevant conclusions and future work in Section~\ref{sec:conclusion}.

% Section System Setup
\section{SYSTEM SETUP}
\label{sec:system_setup}

The system setup consists of two main hardware elements: a robot arm manipulator and one or more depth 3D sensors with a visual camera, in our case Kinect sensors.

With the main goal of achieving an environment-aware robot arm manipulator, the robot is thought to be in the center of the setup with sensors observing it from surrounding angles. Positions of the sensors do not need to be fixed, however, in case one of them is being repositioned, the Eye-to-Hand part of the calibration process has to be repeated.

In the described setup, two Kinect V2 depth sensors were used, observing the robot arm end effector from two viewpoints, each angled at approximately 45{\degree} and one Kinect V1 facing the robot directly. The setup can be seen in Figure \ref{fig:system_setup}. However, the number of sensors is flexible, and only one, or as many as needed can be used as long as sufficient computing power is provided.

\subsection{Calibration Checkerboard}
\label{ssec:calibration_checkerboard}

\begin{figure}[h]
\centering

\begin{subfigure}[t]{0.23\textwidth}
    \includegraphics[width=\textwidth]{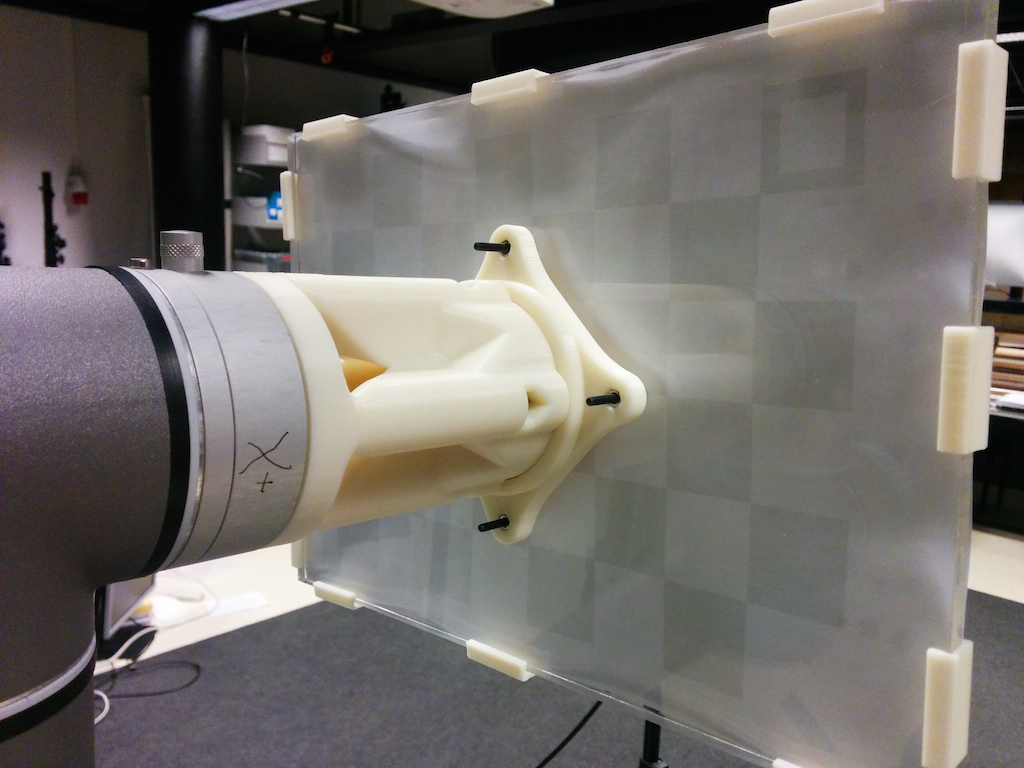}
    \caption{A custom end-effector mount with a rigid plexiglass base for holding a checkerboard.}
    \label{fig:cb_mount}
\end{subfigure}
~
\begin{subfigure}[t]{0.23\textwidth}
    \includegraphics[width=\textwidth]{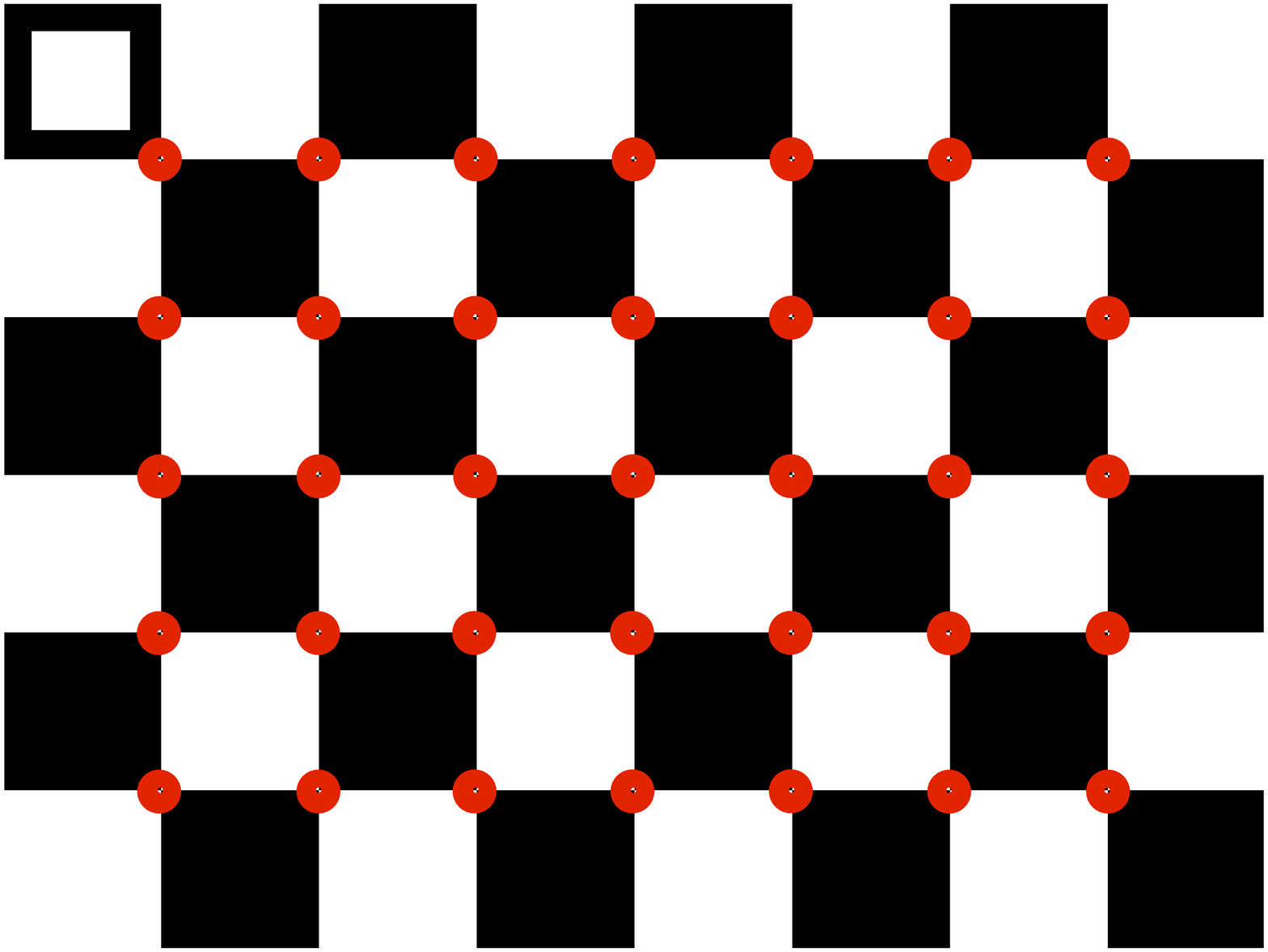}
    \caption{Detected square intersection points are marked in red and a hollow square in the top-left corner, for orientation detection.}
    \label{fig:checkerboard}
\end{subfigure}
\caption{Checkerboard and a custom robot mount.}
\label{fig:checkerboard_on_robot}
\vspace{-0.5cm}
\end{figure}

A custom end-effector mount to hold a checkerboard, with an extension to reduce the number of robot self-collisions, was 3D printed and attached to the end-effector, shown in Figure \ref{fig:cb_mount}. The checkerboard contains 7 by 5 squares, each one of 30 mm by 30 mm size, printed on an A4 paper sheet, which is mounted on hard plexiglass surface to prevent any deformation. One of the side squares is modified to be hollow, as shown in Figure \ref{fig:checkerboard}, and is used to identify correct orientation as described in Section \ref{sec:method}.

\subsection{Robot}

The robotic manipulator being used is UR5 from Universal Robots with \textit{6 degrees of freedom}, a working radius of \textit{850 mm} and a maximum payload of \textit{5 kg}. The repeatability of the robot movements is \textit{0.1 mm}.

\subsection{Sensors}

In our research we include the novel low-cost Kinect V2 sensor~\cite{Fankhauser2015KinectV2ForMobileRobotNavigation}. It has been shown to achieve a significantly higher accuracy compared to its predecessor Kinect V1~\cite{amon2014evaluation}. Kinect V2 is based on \textit{time-of-flight (ToF)} approach, using a different modulation frequency for each camera, thus allowing multiple ToF cameras to observe the same object without any interference~\cite{foix2011lock}. For comparison reasons, and to demonstrate the flexibility of the system, one Kinect V1 sensor is also included in our setup. Table \ref{table:kinect_data} summarises technical specifications of Kinect V1 and V2 sensors. Despite both sensors being named Kinect, they are significantly different, requiring separate drivers and, as it was mentioned, are based on different sensing approaches. In general, any 3D camera, with ROS support, can be used with our system.
%Kinect V2 has $1920$x$1080$ pixel resolution color camera operating at $30$ Hz and $512$x$424$ pixel resolution depth sensor with $0.5$ to $4.5$ \textit{meters} depth range operating at $30$ Hz.

% Kinect specs table
\begin{table}[h]
\caption{Kinect V1 and V2 Technical Specifications.}
\label{table:kinect_data}
\centering
    \begin{tabular}{| l | l | l |}
    \hline
     & \textbf{Kinect V1} & \textbf{Kinect V2} \\ \hline
    \textbf{Sensor type} & Structured Light & Time-of-Flight\\ \hline
    \textbf{RGB Cam Resolution} & $640$x$480$ & $1920$x$1080$\\ \hline
    \textbf{IR Cam Resolution} & $320$x$240$ & $512$x$424$\\ \hline
    \textbf{Refresh Rate} & $30$ Hz & $30$ Hz\\ \hline
    \textbf{Depth Range} & $0.4$ to $4.5$ \textit{meters} & $0.5$ to $4.5$ \textit{meters}\\ \hline
    \textbf{Field of View Horizontal} & 57\degree & 70\degree \\ \hline
    \textbf{Field of View Vertical} & 43\degree & 60\degree \\
    \hline
    \end{tabular}
    \vspace{-0.2cm}
\end{table}

\begin{figure}[h]
\centering
\includegraphics[width=0.48\textwidth]{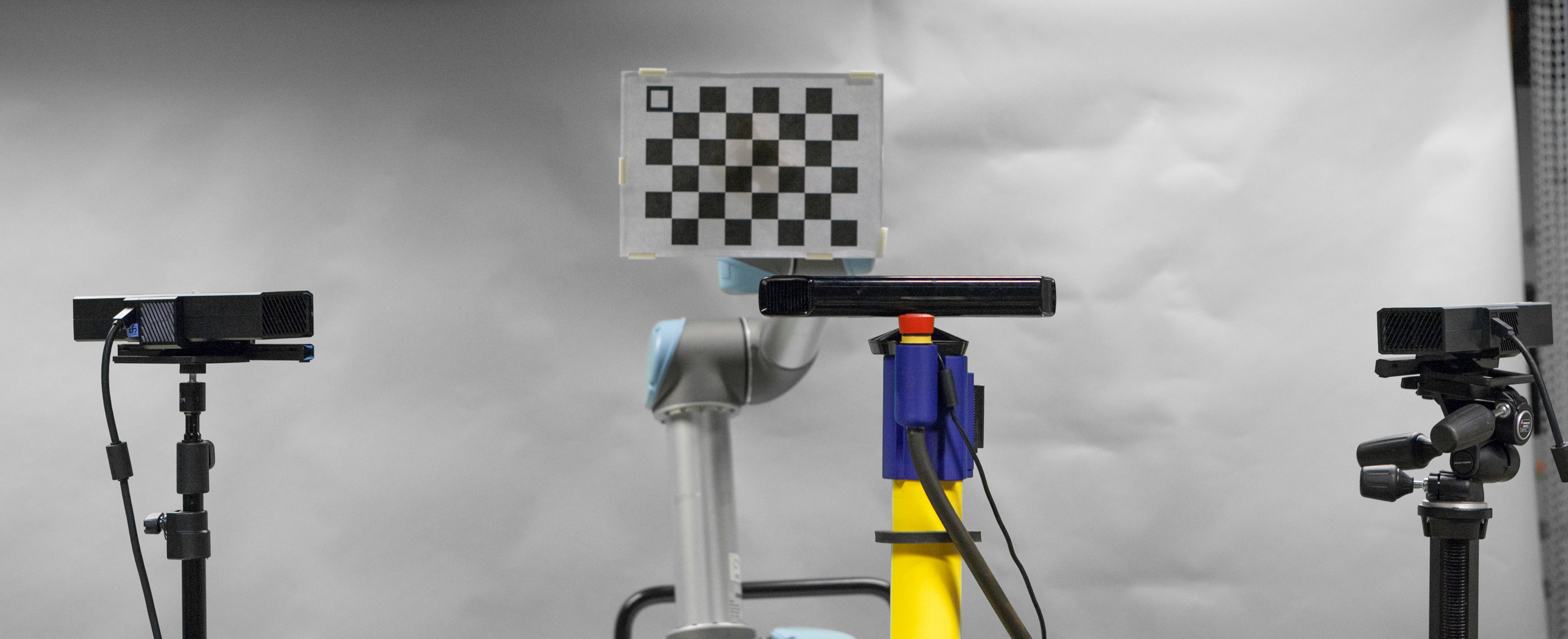}
\caption{Picture of the setup. A checkerboard with a hollow square to allow the detection of its orientation is attached to the robot.}
\label{fig:robot_setup_photo}
\vspace{-0.4cm}
\end{figure}

\subsection{Software}

The whole system software is based on the Robot Operating System (ROS), an open-source meta-operating system running on top of Ubuntu 14.04~\cite{quigley2009ros}. The main advantage of using ROS is its modular design allowing the algorithm to be divided into separate smaller modules performing separate tasks and sharing the results over the network. The workload in our setup was divided over multiple machines, one for each of the 3D cameras and a central one coordinating all the modules and controlling the robot.

Kinect V2 is not officially supported on Linux, however, open-source drivers including a bridge to ROS were found to function well, including the GPU utilisation to improve the processing speed of large amounts of data produced by sensors~\cite{iaikinect2}. Well tested OpenNI 2 drivers were used to integrate Kinect V1 into the system.

The modular design allows for interchanging any of the modules without the need to make any modifications to the rest of the system. For example, any of the depth sensors can be exchanged to another model, or another robotic manipulator can be used, as long as the inter-modular message format is kept the same. Furthermore, addition of extra depth sensors to the system only requires adding an extra message topic for the coordinating module to listen to.

% Section Methods
\section{METHOD}
\label{sec:method}

Our proposed automatic calibration approach consists of a number of modules working together to achieve the desired accuracy of calibration. The calibration can be divided into two main parts:
\begin{enumerate}
  \item Sensor internal parameter calibration
  \item Eye-to-Hand calibration
\end{enumerate}

We first present the general overview of the system functionality and then go into details of each of the processes.

\subsection{Overview of the Whole System Functionality}

\begin{figure}[h]
\centering
\includegraphics[width=0.44\textwidth]{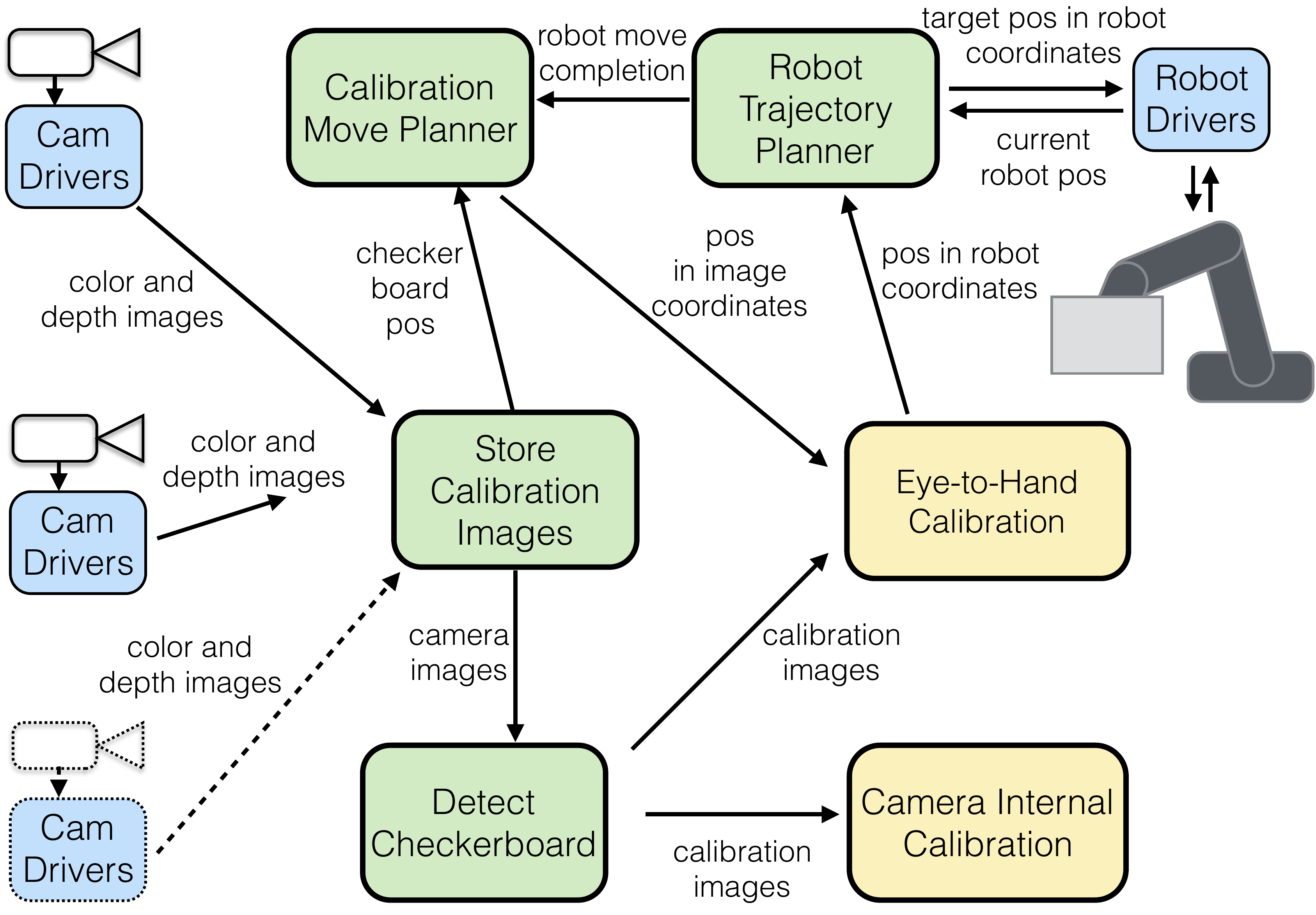}
\caption{The whole framework overview including all the modules and the sequence of the processes. Drivers are marked in blue, image analysis and move planning modules are marked in green and actual calibration modules are marked in yellow. A possibility to add additional 3D cameras to the system is represented by the objects in dashed lines.}
\label{fig:framework_overview}
\vspace{-0.2cm}
\end{figure}

The structure of the whole calibration framework is shown in Figure~\ref{fig:framework_overview}. A specific processing is performed by each module and the information between modules is exchanged using custom messages. Instead of having one central unit, each module publishes messages on defined topics to which other modules can subscribe to, resulting in an asynchronous direct peer-to-peer communication. Each message has a time-stamp to allow synchronization and ignoring out-of-date messages. Updating or interchanging modules can be done even at run time as long as the message format is kept identical. Additional sensors can be added in a similar manner, with the new sensor's message topics, which stream IR and RGB images, added to the configuration file, so that it is seen by the rest of the system. It has to be made sure that each camera uses unique message topic names. An overview of the whole calibration process is presented below. Algorithm~\ref{algo:calibration_process} describes a step-by-step process performed for each camera after the system is launched and $360\degree$ initialization movement is performed.

\begin{algorithm}[h]
\begin{algorithmic}
\STATE{Initial Eye-to-Hand calibration}
\STATE{Tilting motion to define max angles}
\STATE{Estimate the end-effector attachment offset}
\STATE{Generate the robot movement trajectory}
\LOOP
    \STATE{Move the robot to the next position}
    \STATE{Detect checkerboard}
    \IF{detected}
        \STATE{Save images}
        \STATE{Calculate the accumulative Eye-to-Hand calibration}
        \STATE{Apply this calibration}
        \STATE{Recalculate the remaining robot movement trajectory}
    \ENDIF
\ENDLOOP
\STATE{All robot movements are finished}
\STATE{Calculate the internal calibration using saved images}
\STATE{Convert saved images using internal calibration}
\STATE{Calculate the full Eye-to-Hand calibration}
\end{algorithmic}
\caption{Calibration process for each camera}
\label{algo:calibration_process}
\end{algorithm}

\subsection{Checkerboard Detection}

Existing algorithms included in the OpenCV library were used for checkerboard detection in both color and depth data~\cite{bradski2008learning}~\cite{opencvchessboard}. Real-time performance is achieved with X and Y coordinates of identified intersection points of squares on the checkerboard, defined as corners, shown in Figure~\ref{fig:checkerboard}, and depth value obtained from the depth data. Given the noisy depth data, a normalized value from the surrounding area of 10 pixels over 5 consecutive frames is taken and a median value was calculated to reduce the effects of the sensor noise.

Positions in 3D coordinates of the same checkerboard corners are simultaneously calculated using the robot encoder data, corrected with the estimated offset of the checkerboard mounting. Initially, the assumption is made that the checkerboard center point matches the end-effector center point. Then tilting motions of the end effector in place are performed while observing changed positions of the checkerboard corners. Calculation 3D affine transformation and error minimization between expected corner positions and real ones, provides an accurate offset of the end-effector mount. The end-effector mount has to be rigid. Both the data from 3D cameras and from robot encoders are fully synchronised according to the timestamps of when it was captured to reduce any accuracy issues.

Given the four possible orientations of the checkerboard, the modified corner square of the checkerboard, seen in the top left of the checkerboard in Figure~\ref{fig:checkerboard}, is detected using binary thresholding method and the orientation is noted. With the collected data, the corresponding checkerboard corner data points can be matched.

\subsection{Sensor Internal Parameter Calibration}

RGB-D cameras are calibrated for internal parameters using the method proposed by Zhang~\cite{zhang2000flexible} in order to compensate for the following systematic errors:
\begin{enumerate}
  \item Color camera lens distortion
  \item Infrared (IR) camera lens distortion
  \item Reprojection error, or color to depth image offset
  \item Depth distortion
\end{enumerate}

Other non-systematic and random errors like amplitude-related errors or temperature-related errors are not discussed or analysed in this paper, because standard internal camera parameter calibration procedure does not compensate for them, and they are not crucial in current application~\cite{Fankhauser2015KinectV2ForMobileRobotNavigation}~\cite{smisek20133d}.

\subsection{Eye-to-Hand Calibration}

Using the corresponding 3D corner points of the calibration checkerboard, a 3D Affine transformation matrix between the 3D camera and the robot end effector is estimated~\cite{opencvchessboard}. With some likelihood of imprecise detection of checkerboard corners, the outlier detection based on Random Sample Consensus (RANSAC) method is being used on the inputs~\cite{fischler1981random}. The outcome of the estimator is a 3x4 Affine transformation matrix seen in Equation ~\ref{eq:affine_matrix}, where $R$ is a $3x3$ rotation matrix and $t$ is $3x1$ translation vector.

% FIX ERRORS!
\begin{equation}
 T_{C_{3}}^{R} = 
\left \{
  \begin{tabular}{cc}
  $R_{3x3}$ & $t_{3x1}$ \\
  $0_{1x3}$ & $1$
  \end{tabular}
\right \}
\space 
\label{eq:affine_matrix}
\end{equation}

Using the calculated transformation matrix, the 3D points detected in 3D camera color image and depth data can be transformed from the camera coordinate system to the robot's base coordinate system.

\subsection{Robot Motion Planning}

Robot arm control in Cartesian coordinates is used in the project, given the relatively simple movements, as well as limited workspace. Multiple motion planning algorithms included in the \textit{MoveIt!} framework~\cite{sucan2013moveit} were tested. The RRT-connect approach~\cite{kuffner2000rrt}, based on the Rapidly exploring random tree, was found to be suitable for the task. We used an implementation from the from the Open Motion Planning Library (OMPL)~\cite{sucan2012open}.

\begin{figure}[h]
\centering
\includegraphics[width=0.46\textwidth]{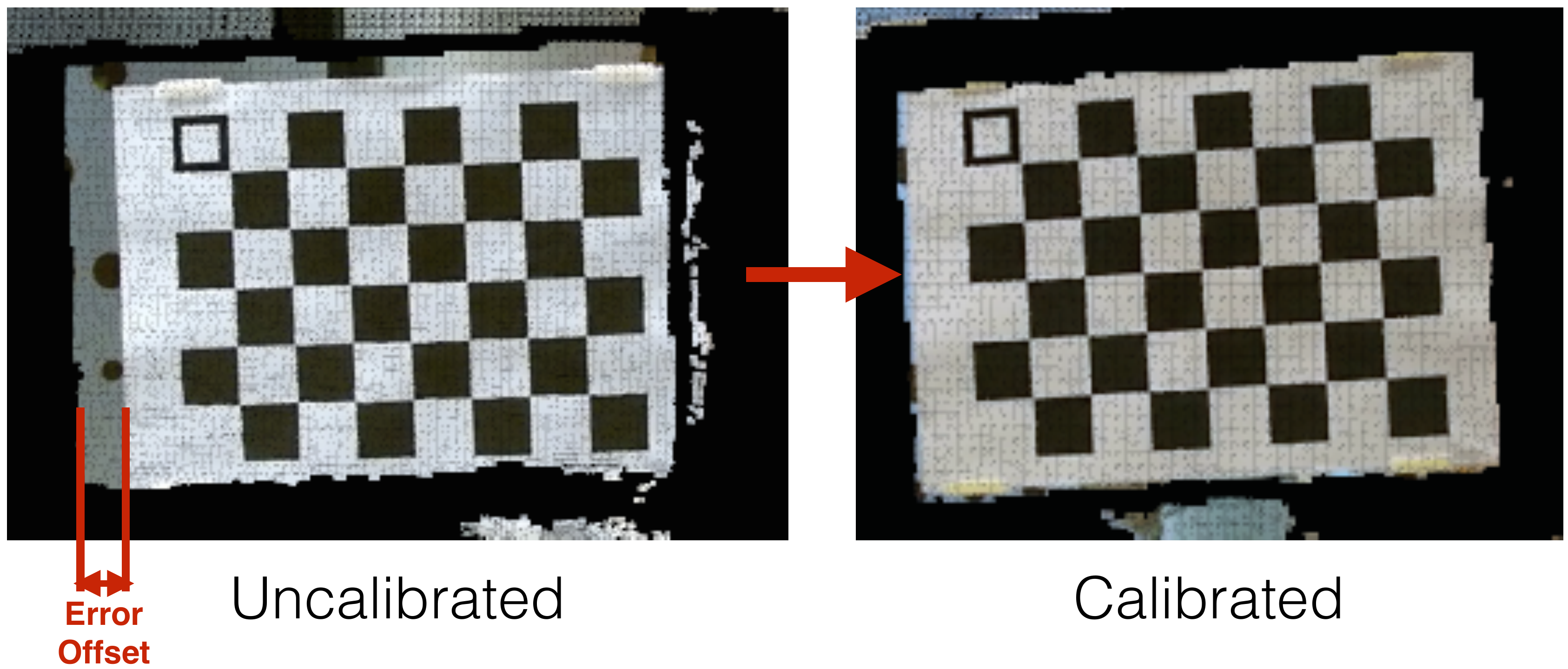}
\caption{Reprojection error shown in the color image and depth point cloud overlay. The offset in the left image is caused by imprecisely defined relative positions between the color and infrared cameras in the 3D camera. Internal camera calibration compensates for this error. The result is seen in the image on the right side, where the offset is reduced.}
\label{fig:reprojection_error}
\vspace{-0.5cm}
\end{figure}

In order to achieve a high-quality 3D camera internal calibration, the samples should include the checkerboard positioned in the majority of the camera's field of view and for least at two distances. Furthermore, tilting the checkerboard at different angles in relation to the camera increases the calibration accuracy~\cite{iaikinect2}. Knowing the calibration pattern parameters, tilting motion allows for more accurate mapping from 3D world coordinates to the 2D image sensor coordinates, based on the projection lines of the known calibration pattern~\cite{zhang2000flexible}.

In order to simplify the internal calibration, it was decided to collect all the data at once and then the calibration using the whole dataset was calculated. However, this meant that lens distortion was still present during the data collection. Furthermore, a reprojection error occurs, which is an offset between the color image and depth data, shown in Figure~\ref{fig:reprojection_error}. These issues caused the Eye-to-Hand transformation to be imprecise, especially for points closer to the edge of the camera image, where lens distortion is more significant.

\begin{figure}[h]
\centering
\includegraphics[width=0.42\textwidth]{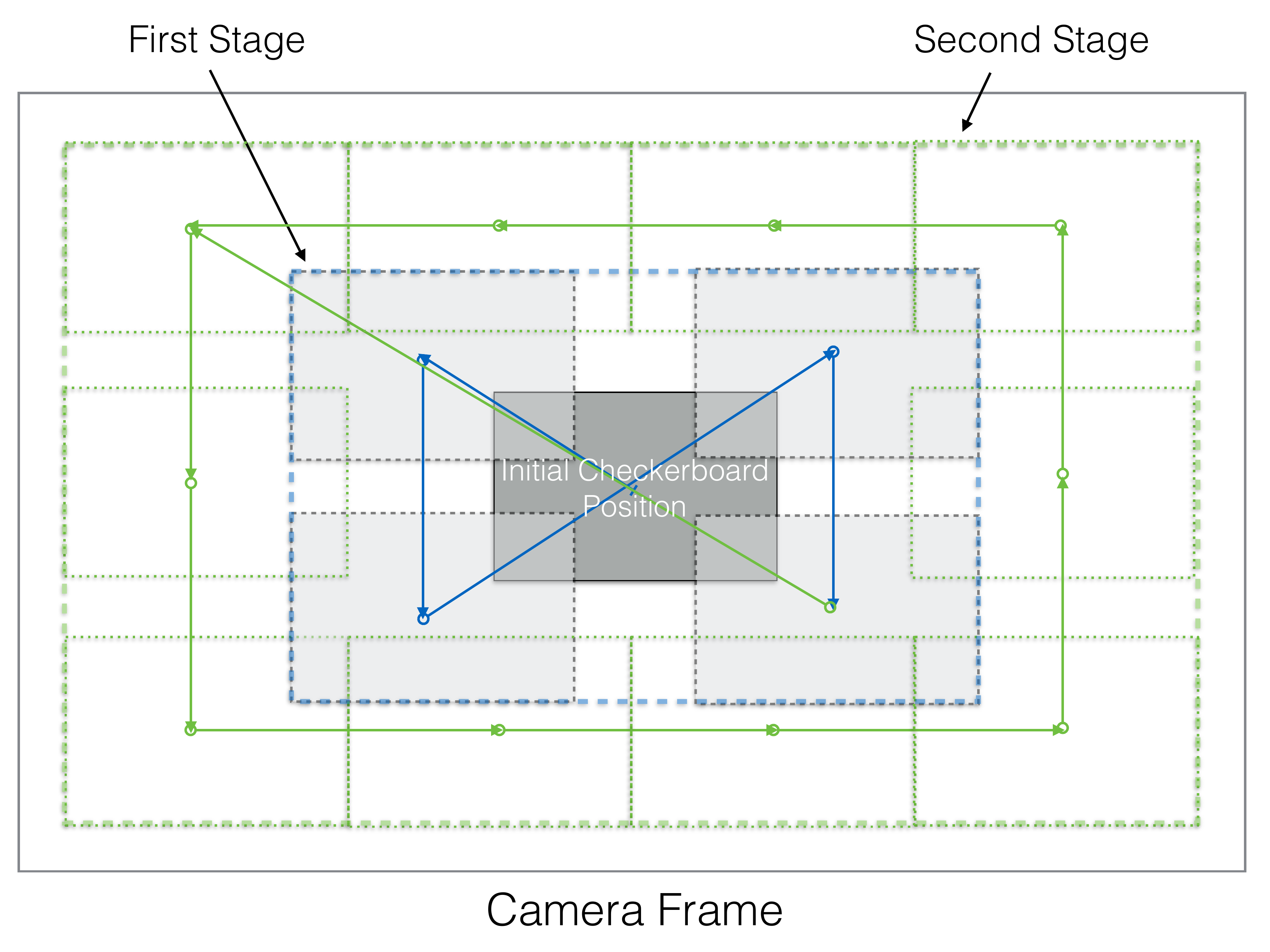}
\caption{Robot movement trajectory as seen in the 3D camera image. It is split into multiple stages by the positions calculated at increasing distance from the center point of the image. Movements are done stage-by-stage, while improving the Eye-to-Hand transformation accuracy at each step. This figure shows just a two stage example.}
\label{fig:robot_move_stages}
\vspace{-0.2cm}
\end{figure}

A robot movement trajectory was chosen with small offsets from the starting point. It can be split into multiple stages by the positions calculated at increasing distance from the center point of the image. The number of stages depend on the overlap level of the checkerboard positions in the camera image, size of the checkerboard, reach-ability of the robot arm as well as the size of the area covered by the camera. At each new position, the detected checkerboard intersection points are accumulated and Eye-to-Hand calibration was recalculated to continually improve the accuracy. The example two-stage robot movement trajectory is shown in Figure \ref{fig:robot_move_stages}. However, this data is not considered for the final Eye-to-Hand calibration, because the 3D camera sensor itself is still not calibrated at this point.

This approach has shown to reduce the robot movement error and by the time positions close to the image edges are chosen, the transformation is accurate enough not to exit the camera's field of view, where checkerboard corners cannot be detected anymore.

During the calibration movements, the checkerboard is tilted by defining changing roll, pitch and yaw of the end effector. During the initialisation of the calibration, a checkerboard is turned to face the 3D sensor directly and then tilted to each direction at 5 degree increments, both positive and negative rotation direction, while trying to detect the checkerboard. Once the detection fails, the tilting is backtracked 
%by $10 \%$ 
and the angle is saved as a maximum allowed tilting. The same process is performed for roll, pitch and yaw to positive and negative angle limits. Roll angle is limited to $\pm45{\degree}$. This angle can vary significantly depending on the type of 3D camera as well as lighting conditions. Reflections caused by the room lights can affect the checkerboard detection.
%ranging from -45{\degree} to 45{\degree} with 45{\degree} steps. 

At each position, images and detected 3D corner positions in color and infrared camera images are saved at each pose. If the desired point is outside the robot workspace, it is automatically identified by the planning algorithm and skipped. Given the positions are reachable, the same trajectory is performed with a $20\%$ higher depth offset away from the camera in order to have data at different distances from the sensor calibrated, while the checkerboard still appears large enough in the image to be detected.

Once the planned movement trajectory is completed, internal 3D camera calibration is performed using the collected data.

\subsection{Repeated Eye-to-Hand Calibration}

After the internal calibration of each 3D camera, the accuracy of the Eye-to-Hand calibration is not precise because of compensated lens distortion and adjustments to reduce the reprojection error. The simplified move sequence, without the tilting, is repeated with the robot moving to previously visited positions by reusing the same coordinates and just Eye-to-Hand calibration recalculated. Once this process is finished, the sensor in the system is fully calibrated.

\subsection{Checkerboard Observation}

As mentioned previously, a flexible number of 3D cameras can be calibrated with the system. Inclusion of the additional sensor into the system is done by defining a configuration file containing the topic names the camera is publishing on. While one 3D camera is being calibrated, any other sensors included in the system are passively observing the robot and running the simplified checkerboard detection algorithm. If the checkerboard is detected, the pose of the checkerboard and the pose of the robot, which is being streamed on the network by the robot controlling node, are recorded. In any subsequent checkerboard detection instances, the position is compared to the position of the previous detection, and if the current one is closer to the center of the color image, the poses are updated in order to have a more reliable starting position. Once the current calibration of a 3D camera is completed, the request is sent to the robot to move to the detected position and start the calibration procedure for the other sensor.

% Section Experiments and Results
\section{Experiments and Results}
\label{sec:experiments}

%\textcolor{red}{
%Experiments to be performed
%\begin{itemize}
%  \item Accuracy: uncalibrated system vs calibrated system vs robot encoders vs motion capture
%  \item Number of recorded poses vs the calibration accuracy
%  \item Different tilting at each position vs calibration accuracy
%  \item Calibration estimation for the specific points in the camera image and improving only in the needed positions
%\end{itemize}}

%\subsection{Calibration Time}

%Text here.

%\subsection{Calibration Accuracy}

%Text here.

The presented calibration process was successfully performed provided that the checkerboard was detected by the 3D camera to be calibrated. Initially, the robot was placed in an upside-down "L" shaped joint configuration and turned $360\degree$ to increase the chances of the checkerboard being detected by the 3D sensors. However, this movement has to be supervised by a human operator to avoid hitting any obstacles. In other cases, the robot was repositioned manually to make sure that the checkerboard was within the field of view of the camera.

Given a close to autonomous operation of our framework, we conducted experiments to analyze the number of checkerboard positions recorded versus the achieved calibration accuracy. As the process for the internal sensor calibration is identical for each of the 3D cameras, for easier comparison, the results from one Kinect V2 camera is presented in the experiments section. Meanwhile, the Eye-to-Hand calibration results have been acquired using the setup described in Section~\ref{sec:system_setup}. Results are divided into two sections according to the calibration type, each one requiring an independent set of robot moves and data collection:
\begin{enumerate}
  \item Internal camera calibration
  \item Eye-to-Hand calibration
\end{enumerate}

\subsection{Internal Camera Calibration}

The first iteration of movements was made in order to calibrate the 3D camera internally, using the robot trajectory explained in Figure~\ref{fig:robot_move_stages}. Because the field of view of the color camera and the infrared camera in the sensor are different, the checkerboard was not always visible or successfully detected in both cameras at the same time. This explains the varying number of detections in each of the sensor's cameras, as well as simultaneously in both, which we refer to as \textit{combined}. There were 9 experiments conducted in total. Experiments 1-2 had large overlap in checkerboard positions and tilting, experiments 3-6 had no overlap anymore and experiments 7-9, no more tilting. Experiment data is summarized in Table~\ref{table:internal_calib_summary}.

\begin{table}[h]
\caption{Experiment data for internal Kinect V2 sensor calibration.}
\label{table:internal_calib_summary}
\centering
%\begin{tabular}{ | c || c | c | c | c | }
\begin{tabular}{ |p{0.7cm}||p{0.75cm}|p{0.75cm}|p{1.1cm}|p{0.8cm}|p{0.65cm}|p{0.8cm}|}
 \hline
 Exp \# & Color Frames & IR Frames & Combined Frames & Overlap & Tilting & Time (sec)\\
 \hline
 Exp 1 & 234 & 215 & 158 & Yes & Yes & 613\\
 Exp 2 & 120 & 109 & 81 & Yes & Yes & 338\\
 Exp 3 & 78 & 72 & 55 & No & Yes & 218\\
 Exp 4 & 57 & 54 & 45 & No & Yes & 176\\
 Exp 5 & 44 & 41 & 33 & No & Yes & 142\\
 Exp 6 & 39 & 35 & 26 & No & Yes & 128\\
 Exp 7 & 15 & 14 & 14 & No & No & 57\\
 Exp 8 & 10 & 9 & 7 & No & No & 45\\
 Exp 9 & 5 & 5 & 5 & No & No & 36\\
 \hline
\end{tabular}
\vspace{-0.2cm}
\end{table}

\begin{figure}[h]
\centering
\includegraphics[width=0.48\textwidth]{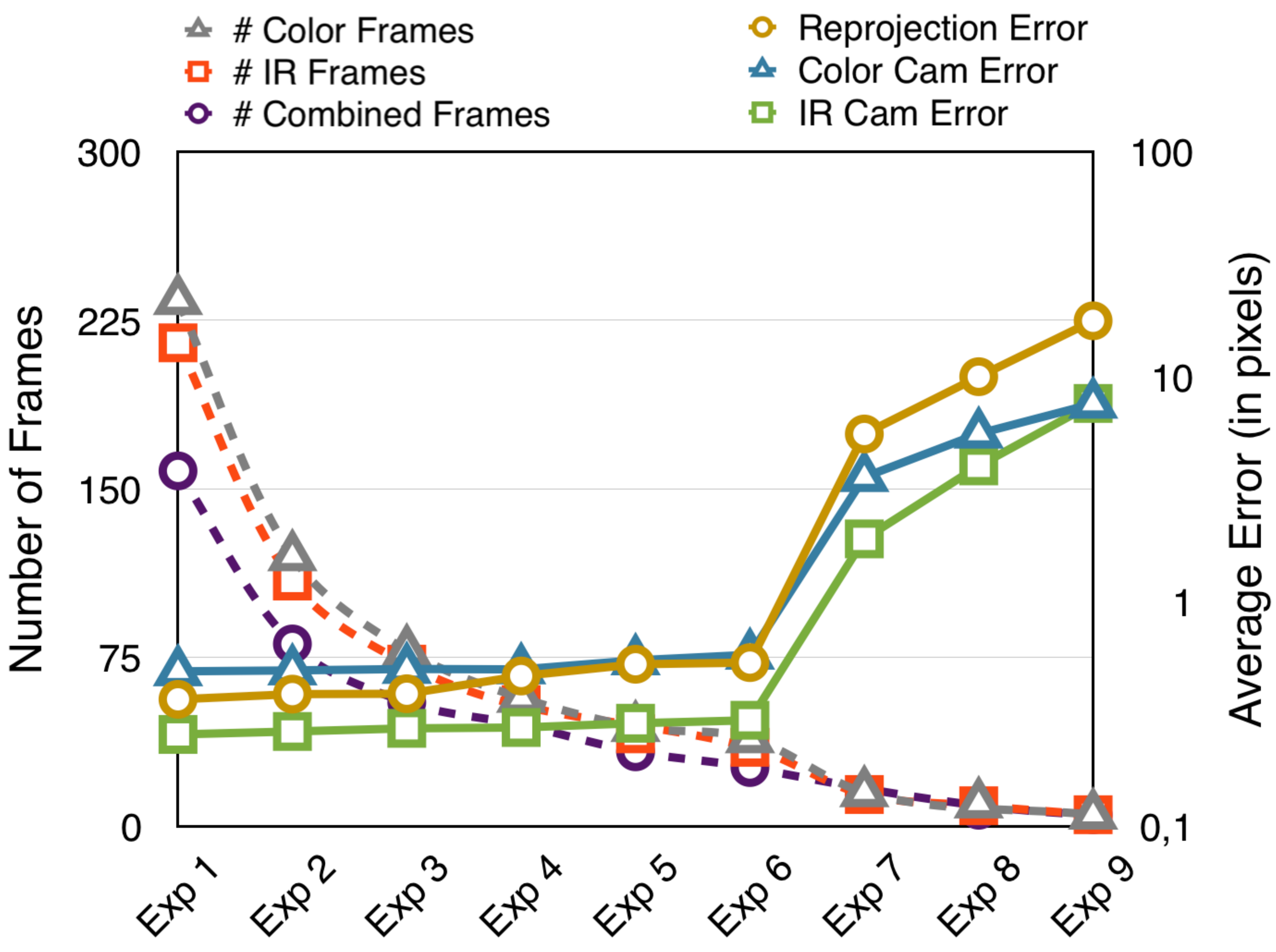}
\caption{Calibration accuracy results by showing the errors of internal 3D camera calibration. Color camera, IR errors define averages of each sensor's cameras. Reprojection error defines the average error of the offset in the images between the color image and the depth information, seen in Figure~\ref{fig:reprojection_error}. It has to be noted that right Y axis for error rates is in log scale.}
\label{fig:results_chart}
\vspace{-0.2cm}
\end{figure}

Figure~\ref{fig:results_chart} shows the calibration results by analyzing the average error in pixels of each of the sensor's cameras and the reprojection error for each of the experiments. Errors were calculated using the known geometry and size of the checkerboard and comparing the calibrated sensor estimation of the checkerboard dimensions according to the square intersection points to the known geometry. The higher the error, the lower the calibration accuracy. 

\subsection{Eye-to-Hand Calibration}

\begin{table}[h]
\caption{Experiment data for Eye-to-Hand calibration.}
\label{table:eyehand_calib_summary}
\centering
%\begin{tabular}{ | c || c | c | c | c | }
\begin{tabular}{ |p{0.7cm}||p{1.0cm}|p{1.0cm}|p{1.0cm}|p{1.0cm}|p{1.2cm}|p{1.9cm}|}
 \hline
 Exp \# & Frames Cam 1 & Frames Cam 2 & Frames Cam 3 & Overlap & Time (sec) \\
 \hline
 Exp 1 & 82 & 80 & 71 & Yes & 470\\
 Exp 2 & 44 & 45 & 39 & Yes & 243\\
 Exp 3 & 14 & 14 & 11 & No & 115\\
 Exp 4 & 9 & 10 & 8 & No & 85\\
 Exp 5 & 5 & 6 & 5 & No & 60\\
 \hline
\end{tabular}
%\vspace{-0.2cm}
\end{table}

The second iteration of moves were performed for Eye-to-Hand calibration, while using the most accurate internal 3D camera calibration mentioned previously. In this part, tilting was not performed and the calibration checkerboard was kept at a constant angle, parallel to the each camera's image plane. 5 experiments using 3 cameras were conducted in total, each using a different number of frames, as seen in Table~\ref{table:eyehand_calib_summary}. Experiment 1 had a large overlap in checkerboard positions, in experiment 2 there was a small overlap, while in the rest there was no overlap and even some gaps between the positions. Cam 1 and 2 were Kinect V2 sensors, while Cam 3 was Kinect V1. Time was measured from the start to the final calibration result of all three cameras.

\subsection{Result Analysis}

For the internal calibration, it can be seen that in the first 6 experiments, even with a significantly lower number of frames used, the error in all of the sensor's cameras did not increase much. However, experiments 7 to 9, where the calibration checkerboard was present only in the part of the camera's field of view and was not tilted, show a significant increase in errors. It can be concluded that the most important part to achieve good internal calibration accuracy is to cover the field of view of the camera and perform tilting, but overlapping same areas with the checkerboard is not mandatory.

\begin{figure}[h]
\centering
\includegraphics[width=0.48\textwidth]{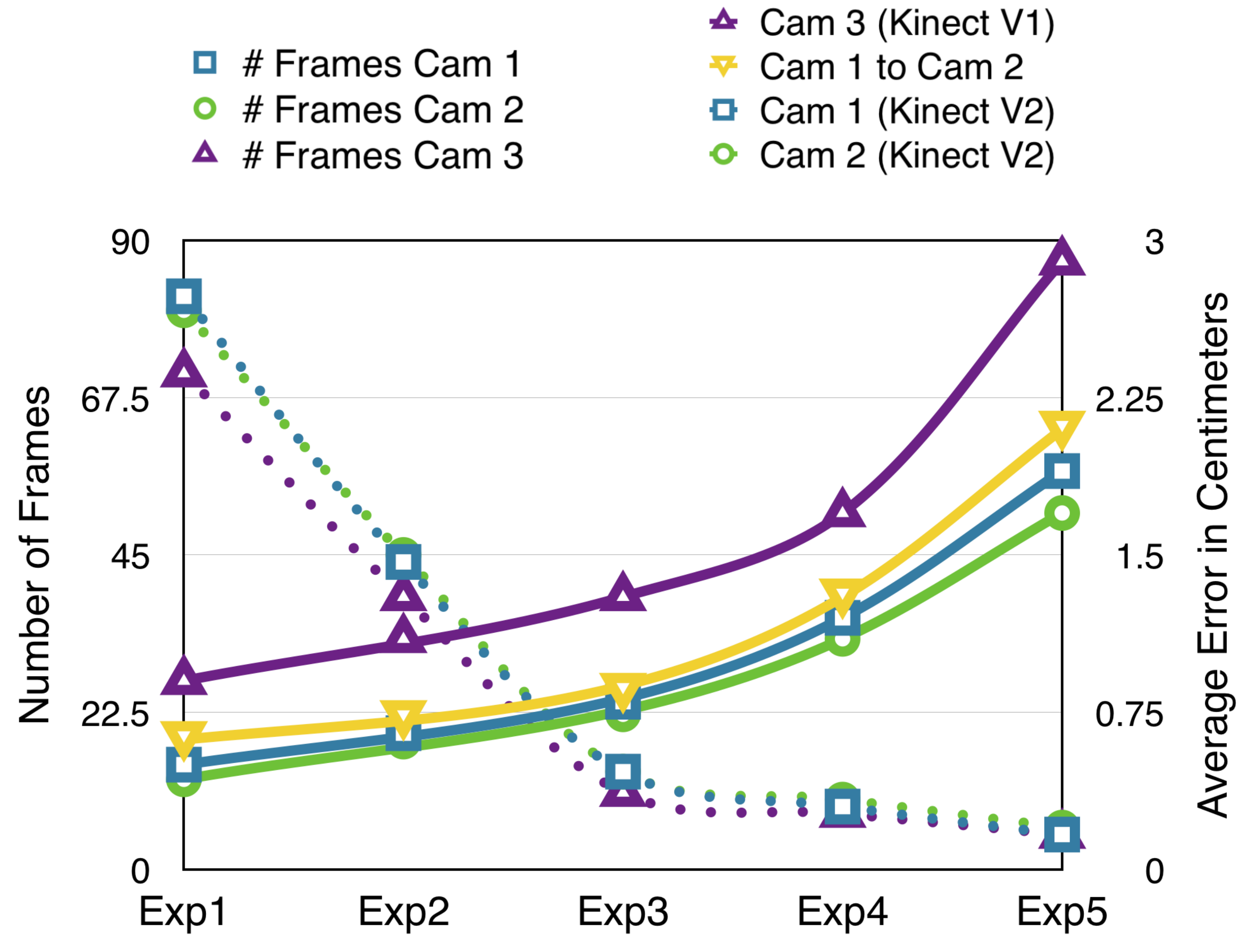}
\caption{Eye-to-Hand calibration accuracy results. Overall position error (in cm) as well as each axis separately are shown by comparing the actual robot position versus the predicted robot position from the 3D camera sensor. Dotted lines indicate the number of frames used in each experiment.}
\label{fig:results_chart2}
\vspace{-0.2cm}
\end{figure}

Figure~\ref{fig:results_chart2} shows that the average error rate of Eye-to-Hand calibration has inverse correlation to the number of frames used. The larger area of the camera's field of view is used, the more accurate calibration is achieved. Looking at the overall average error, experiment 3 seems to be the most optimal choice for all three cameras when considering the number of frames used and accuracy achieved.

As expected, the older Kinect V1 was significantly less accurate compared to Kinect V2, mainly due to lower resolution RGB and IR sensors in the camera. The average error between camera 1 and camera 2, both being Kinect V2, was almost the same as average errors of each sensor to the robot. Therefore, calibration of each 3D camera to the robot is enough for joining the point clouds of two cameras, and no additional calibration is necessary.

%Interesting effect of Z-axis having a significantly larger error compared to X and Y was observed. 
It was also  noticed in all five experiments that the Z-axis has on average $20\%$ larger error compared to both X and Y-axis. It is likely to be caused by the noisy depth data from 3D camera, which should be compensated using more specific methods. On the other hand, it proves that position estimation in X and Y-axes has even lower error than our indicated overall error of the calibration.

\section{CONCLUSION AND FUTURE WORK}
\label{sec:conclusion}

A simple and flexible calibration method for systems containing a robot and one or more 3D cameras was presented. It is based on the robot moving a standard calibration checkerboard and being guided by the information sent from each of the cameras to cover the largest possible area in the field of view, to ensure an accurate calibration.

A full calibration, including a sensor internal calibration together with an Eye-to-Hand calibration can be done, or just the second part separately, given that the sensor is already calibrated internally. Modular design ensures that sensors can be added or removed easily, as well as hardware components interchanged without any modifications to the algorithm.

According to experiment results, achieving good calibration requires the robot to cover the majority part of the field of view of the 3D camera to achieve a good accuracy. Using our system, a good accuracy calibration of one 3D sensor taken just out of the box, can be achieved in just a few minutes with minimal supervision by the operator. This makes the system integration and reconfiguration significantly faster compared to standard manual method, while keeping the flexibility of varying configurations. 

An example application the calibration process was aimed at the environment-aware collaborative robot arm, where people or other moving objects can freely and safely operate in the workspace of the robot without a risk of collision. However, for a more precise operation where sub-centimeter accuracy is necessary, a different, more up-close, setup would be needed as well as a checkerboard containing a finer structure.

The framework will be further tested with a variety of physical setups and different 3D cameras and multiple robot arm types. We plan to open source the code, making it accessible to researchers allowing further testing and development.

Algorithm improvements will include a simultaneous calibration of multiple cameras provided that the calibration checkerboard is within the field of view. Furthermore, if only part of the field of view of the camera will be used in the operation, it could be defined by the user and instead of calibrating the whole image area, only the area of interest would be used. 
%
%Automatic detection of the existing calibration validity will be added. A number of robot moves, guided by the camera, will be made to ensure that the end effector stays within the field of view. Then, the positions from robot joint encoders will be compared to the estimation of the camera and error measurements used to determine whether the calibration is still accurate enough.

\addtolength{\textheight}{-12cm}   % This command serves to balance the column lengths

%%%%%%%%%%%%%%%%%%%%%%%%%%%%%%%%%%%%%%%%%%%%%%%%%%%%%%%%%%%%%%%%%%%%%%%%%%%%%%%%

%%%%%%%%%%%%%%%%%%%%%%%%%%%%%%%%%%%%%%%%%%%%%%%%%%%%%%%%%%%%%%%%%%%%%%%%%%%%%%%%

%%%%%%%%%%%%%%%%%%%%%%%%%%%%%%%%%%%%%%%%%%%%%%%%%%%%%%%%%%%%%%%%%%%%%%%%%%%%%%%%
%\section*{APPENDIX}

\section*{ACKNOWLEDGMENT}
This work is partially supported by The Research Council of Norway as a part of the Engineering Predictability with Embodied Cognition (EPEC) project, under grant agreement 240862

%%%%%%%%%%%%%%%%%%%%%%%%%%%%%%%%%%%%%%%%%%%%%%%%%%%%%%%%%%%%%%%%%%%%%%%%%%%%%%%%

\bibliographystyle{IEEEtran}
\bibliography{IEEEexample}

\end{document}